%% file: neurips_2020.tex
\documentclass{article}

\PassOptionsToPackage{numbers, compress}{natbib}

\usepackage[preprint]{neurips_2020}

\usepackage[utf8]{inputenc} 
\usepackage[T1]{fontenc}    
\usepackage{hyperref}       
\usepackage{url}            
\usepackage{booktabs}       
\usepackage{amsfonts}       
\usepackage{nicefrac}       
\usepackage{microtype}      
\input{math_commands}

\usepackage{graphicx}
\usepackage{multirow}
\usepackage{amssymb}
\usepackage{amsmath}
\usepackage{relsize}
\usepackage{caption}
\usepackage{floatrow}
\usepackage{subfig}
\usepackage{float}
\floatstyle{plaintop}
\restylefloat{table}
\usepackage{array}
\usepackage{siunitx}
\usepackage{tikz}
\usepackage{todonotes}
\usepackage{xcolor}
\newcommand{\cut}[1]{}

\usetikzlibrary{decorations.pathreplacing}
\usepackage{pgfplots}
\usetikzlibrary{intersections, pgfplots.fillbetween}

\title{Learning the Prediction Distribution for Semi-Supervised Learning with Normalising Flows}

\author{%
  Ivana Balažević\thanks{Equal contribution} 
  \hspace{1.2cm} 
  Carl Allen\textsuperscript{*} 
  \hspace{1.cm} Timothy Hospedales
  \\
  University of Edinburgh \qquad Samsung AI Centre, Cambridge \\
  \texttt{\{ivana.balazevic, carl.allen,
  t.hospedales\}@ed.ac.uk}}

\begin{document}

\maketitle

\begin{abstract}

As data volumes continue to grow, the labelling process increasingly becomes a bottleneck, creating demand for methods that leverage information from unlabelled data. Impressive results have been achieved in semi-supervised learning (SSL) for image classification, nearing fully supervised performance, with only a fraction of the data labelled. In this work, we propose a probabilistically principled general approach to SSL that considers the distribution over label predictions, for labels of different complexity, from ``one-hot'' vectors to binary vectors and images. Our method regularises an underlying supervised model, using a normalising flow that learns the posterior distribution over predictions for labelled data, to serve as a prior over the predictions on unlabelled data. We demonstrate the general applicability of this approach on a range of computer vision tasks with varying output complexity: classification, attribute prediction and image-to-image translation.
\end{abstract}

\section{Introduction}
\label{introduction}
Recent years have seen a tremendous increase in the quantity of available data, but the data labelling process remains a costly bottleneck, often requiring much human effort. This issue has motivated the development of machine learning approaches such as zero- or few-shot learning, that are able to learn from small amounts of labelled data by harnessing information from other data; and  \textit{semi-supervised learning (SSL)}, which leverages additional, typically much larger quantities of unlabelled data.

For unlabelled data to be useful in predicting labels, the distribution $p(x)$ over labelled and unlabelled inputs $x \!=\! \{x^l, x^u\}$ must contain information relevant for the prediction \cite{chapelle2006semi}. Many SSL methods rely, often implicitly, on direct assumptions about $p(x)$; others relate to the distribution $p(\theta)$, where each instance of $\theta$, as predicted by e.g. a neural network, \textit{parameterises the distribution} $p(y|x;\theta)$ for a given $x$.
Such assumptions are usually encoded in an unsupervised loss component, added to a supervised loss function of the labelled data, which can be interpreted as regularising the supervised model, encouraging generalisation to unseen data. 
Approaches relying on assumptions about the input distribution $p(x)$ include consistency regularisation \cite{tarvainen2017mean,miyato2018virtual}, where a model is penalised if realistic perturbations of unlabelled data $x^u$ change their predicted label. Methods relying on assumptions about the distribution $p(\theta)$ over predicted parameters include entropy minimisation \cite{grandvalet2005semi}, where a model is encouraged to output ``confident'' predictions on unlabelled data. State-of-the-art holistic approaches include methods of both types \cite{berthelot2019mixmatch, berthelot2020remixmatch}. In this work, we focus on SSL methods that rely on properties of $p(\theta)$, for tasks in which $y|x$ is deterministic. 
Although inputs are, by definition, more abundant than labels, SSL methods relying on $p(\theta)$ are of interest since the complexity of $p(\theta)$ is often far less than that of $p(x)$, in some cases allowing $p(\theta)$ to be described in analytic form. Furthermore, the two options are orthogonal, achieving state-of-the-art results when combined \cite{berthelot2019mixmatch, berthelot2020remixmatch}.

Existing SSL approaches are commonly \textit{task-specific}, e.g. image classification \cite[e.g.][]{kingma2014semi, tarvainen2017mean,miyato2018virtual, nalisnick2019hybrid, berthelot2019mixmatch} or depth estimation \cite[e.g.][]{kuznietsov2017semi, cho2019large}. Impressive results can be achieved by leveraging task-specific assumptions about $p(x)$ or $p(\theta)$, e.g. SSL for image classification with very few labelled data is approaching the upper bound of supervised performance on the whole dataset  \cite{berthelot2019mixmatch, berthelot2020remixmatch}. 
Of relevance to our work, a probabilistic model has recently been proposed for SSL methods that rely on properties of the predicted parameter distribution $p(\theta)$ \cite{allen2020probabilistic}. The model justifies several existing approaches, e.g. entropy minimisation \cite{grandvalet2005semi} and mutual exclusivity \cite{sajjadi2016mutual, xu2018semantic}, as approximating a continuous relaxation of the discrete distribution $p(\theta)$, under the assumption that $y|x$ is deterministic. 
Taking inspiration from this, we propose a \textit{general} approach for semi-supervised learning that relies on properties of $p(\theta)$. 
In our proposed method, we exploit recent advances in density estimation, using \textit{normalising flows} \cite{dinh2017density}, specifically \textit{neural spline flows} \cite{durkan2019neural}, to learn the posterior distribution over the predictions for labelled data, that can be used as a prior over the predictions for unlabelled data.

Similar to entropy minimisation \cite{grandvalet2005semi} and mutual exclusivity \cite{sajjadi2016mutual, xu2018semantic}, our \textit{learned prior (LP)} method is complementary to SSL approaches relating to $p(x)$ and can be used as a component of a holistic model. However, our approach is more general than existing methods relating to $p(\theta)$, requiring no prior knowledge of the prediction distribution. It is thus applicable to tasks with any output type, from classes to multiple binary attributes, even images. Experiments show that our proposed approach performs comparably to methods that embed prior assumptions of $p(\theta)$ on image classification and attribute prediction tasks, but can also be applied to tasks with a more complex output distribution, such as image-to-image translation.

\vspace{-0.1cm}
\section{Related work}
\label{rel_work}

\textbf{Notation:} $x^l_i \!\in\! \mX^l, y^l_i \!\in\! \mY^l$, $i\!\in\!\{1\,...\,N_l\}$ are labelled data points;
$x^u_j \!\in\! \mX^u, y^u_j \!\in\! \mY^u$, $j\!\in\!\{1\,...\,N_u\}$ are unlabelled data points and their respective labels; $\mathcal{X}, \mathcal{Y}$ are domains of $x$ and $y$. 
$\theta$ denotes parameters of the distribution $p(y|x;\theta)$, not weights of a neural network. As a random variable, $\theta$ varies with $x$ under a distribution $p(\theta)$ and $\theta^{x}$ denotes a realisation corresponding to a specific $x$. Thus $\theta$ defines a distribution, and $p(\theta)$ is a distribution over those distributions. 

\vspace{-0.1cm}
\subsection{Semi-supervised learning}
\label{sec:ssl}
\vspace{-0.1cm}

An archetypal loss function for semi-supervised learning  consists of two parts: a standard supervised loss component on the labelled data $\ell_l$ and a weighted component for the unlabelled data $\ell_u$:
\begin{equation}
    \ell_{\text{SSL}} = \ell_{l} + \lambda \ell_{u},  
\end{equation}
where $\lambda$ controls the contribution of the unlabelled term to the overall loss. The labelled loss $\ell_l$ is typically task-dependent, e.g. softmax cross entropy in the case of classification.  
Much existing work on semi-supervised learning for image classification can be broadly categorised according to the type of assumptions underlying the unlabelled loss component $\ell_u$. We categorise them as (i) input distribution ($p(x)$) loss, (ii) prediction distribution ($p(\theta)$) loss, or (iii) holistic, i.e. combining both. 

\textbf{Input distribution loss}\hspace{0.1in}Many existing approaches to SSL rely on the assumed properties of the input distribution $p(x)$. \textit{Data augmentation} refers to techniques that make specific changes to an input image, e.g. cropping, adding noise, that leave its label unchanged. \textit{Consistency regularisation} \cite{sajjadi2016regularization, laine2017temporal,tarvainen2017mean, miyato2018virtual} takes this a step further on the premise that a classifier should output the same class for unlabelled images after augmentation.
Both require domain knowledge of $p(x|y)$ to know how $x$ can be changed such that $y$ remains the same. Extending such techniques to other domains, e.g. sound classification or language modelling, or to computer vision tasks with a more complex output distribution, e.g. semantic segmentation, is typically not  straightforward. 

\textbf{Prediction distribution loss}\hspace{0.1in}Several SSL approaches encourage certain properties in the predictions $\theta^{x^u}$.
\textit{Entropy minimisation} \cite{grandvalet2005semi} encourages the model to make ``confident'' (low-entropy) predictions by the loss term $\ell_u^{\,\text{MinEnt}}(\theta^{x^u}) \!=\! - \frac{1}{N_u}\sum_{j=1}^{N_u}\sum_{k} \theta^{x^u_j}_k\log \theta^{x^u_j}_k$, where $\theta^{x^u_j}_k\!=\!p(y\!=\!k|x)$ for class $k$.
\textit{Mutual exclusivity} \cite{sajjadi2016mutual} encourages class predictions to not overlap through the logical constraint-based loss term $\ell_u^{\,\text{MutExc}}(\theta^{x^u}) \!=\! - \frac{1}{N_u}\sum_{j=1}^{N_u} \log \sum_{y \in \sY} \prod_{k}(\theta^{x^u_j}_k)^{y_k} (1-\theta^{x^u_j}_{k})^{1-y_k}$, where $\sY$ defines the set of ``one-hot'' vectors. \textit{Semantic loss} \cite{xu2018semantic} generalises this to $\sY$ being any set of valid labels. 
In assuming that correct predictions are confident or that classes are mutually exclusive, the above mentioned methods make a priori assumptions about $p(\theta)$ that correspond to $y|x$ being deterministic \cite{allen2020probabilistic}. Specifically, in $K$-class classification, the distribution $p(\theta)$ is described by a weighted sum of point masses at the vertices of the simplex, $\smash{\theta \!\in\! \Delta^K \!\subseteq\!\sR^K}$, where $\smash{\theta^{x}_k \!=\! p(y \!=\!k|x)}, k \!\in\! 
\{1\, ...\, K\}$ and the unlabelled loss function components $\ell_u^\text{\,MinEnt}$ and  $\ell_u^\text{\,MutExc}$ have local maxima at the vertices of the simplex $\Delta^K$ and therefore approximate continuous relaxations of the discrete  $p(\theta)$.

\textbf{Holistic approaches}\hspace{0.1in}MixMatch \cite{berthelot2019mixmatch} and ReMixMatch \cite{berthelot2020remixmatch} are recent state-of-the-art methods on SSL for image classification that combine several techniques, pertaining to both input and output distributions $p(x)$ and $p(\theta)$, achieving impressive results with very small amounts of labelled data. Importantly to our work, advances in each individual component of these holistic approaches are orthogonal and complementary, and so we focus only on SSL methods that relate to $p(\theta)$.

\subsection{Probabilistic framework for discriminative SSL}\label{theory}

A probabilistic model has recently been introduced for \textit{discriminative} semi-supervised learning \cite{allen2020probabilistic}, which explains and unifies entropy minimisation \cite{grandvalet2005semi} and mutual exclusivity \cite{sajjadi2016mutual, xu2018semantic} as approximating continuous relaxations of the distribution over predicted parameters $p(\theta)$. The discriminative SSL model is defined as:
\begin{align}
    p( \mY^l| \mX^l, \mX^u; \omega) = 
    \int_{\alpha} p(\alpha)
        p(\mY^l| \Tilde\theta^{\mX^l})     
        p(\Tilde\theta^{\mX^l}| \alpha)
        p(\Tilde\theta^{\mX^u}| \alpha),
    \label{eq:SSL_new}
\end{align}
where $\theta^{\mX^l\!\!} \!=\! \{\theta^{x^l}\}_{x^l\in\mX^l}$ and $\theta^{\mX^u\!\!} \!=\! \{\theta^{x^u}\}_{x^u\in\mX^u}$ are parameters of distributions $p(y^l|x^l)$ and   $p(y^u|x^u)$ respectively, for all $x^l\!\in\! \mX^l, x^u \!\in\! \mX^u$;
$\alpha$ are parameters of a distribution over $\theta$ (considered a latent random variable);
and $\omega$ parameterises a deterministic function $f_\omega\!:\!\mathcal{X}\!\to\!\Theta$, e.g. a neural network, such that $f_\omega(x) \!\doteq\! \smash{\Tilde{\theta}^{x} \!\approx\! \theta^x}$.

Considering components of Equation \ref{eq:SSL_new}: $p(\mY^l |\, \Tilde\theta^{\mX^l})$ encourages labelled predictions $\Tilde\theta^{x^l}$ to approximate parameters of $p(y^l\,|\,x^l)$; $ p(\Tilde\theta^{\mX^l}| \alpha)$ enables $\alpha$ to capture the empirical distribution over $\Tilde\theta^{x^l}\!$; and $p(\Tilde\theta^{\mX^u}| \alpha)$ causes predictions $\Tilde\theta^{x^u}$ for unlabelled $x^u \!\in\! \mX^u$, to fit the distribution defined by $\alpha$ \cite{allen2020probabilistic}. We re-emphasise that the distribution $p(\theta|\alpha)$, which can be seen to regularise an underlying supervised model, is over model \textit{outputs}, in contrast to the typical case of regularising the model weights $\omega$ (e.g. $\ell_1$/$\ell_2$ regularisation). The distribution $p(\theta|\alpha)$ may be \textit{known} a priori, as implicitly the case in entropy minimisation and mutual exclusivity, or \textit{learned} from labelled data, as in this work.

\subsection{Normalising flows}

Normalising flows \cite{dinh2017density} are generative models for probability density estimation and generation, defined as an invertible mapping $g_\phi: \gX \to \gZ$ from a data space $\gX$ to a latent space $\gZ$. The mapping $g_\phi$ is parameterised by a neural network with an invertible architecture. A distribution over latent variables $p_{\gZ}$ is often chosen to be simple, e.g. a multivariate Gaussian.
The density of the transformed variable $x = g_\phi^{-1}(z)$ is obtained by the change of variables $p_\mathbf{\gX}(x) = p_{\gZ}(g_\phi(x)) \cdot |\det(\partial g_\phi(x)/\partial(x)|$. For a detailed overview of the normalising flows literature, see \cite{papamakarios2019normalizing}. In this work, we use a recent state-of-the-art model, neural spline flows \cite{durkan2019neural}, for learning the prediction distribution $p(\theta)$.

\section{Semi-supervised learning by learning the prediction distribution}

Since, by definition, inputs $x$ are  more abundant than labels $y$ in SSL, it seems natural to leverage information from observed $x$ rather than the predicted estimates $\tilde\theta^x$, i.e. to  rely on the properties of the input distribution $p(x)$. However, that requires task-specific knowledge or assumptions about $p(x)$ that we do not make. Additionally, model outputs often have a far simpler structure than the inputs, e.g. one-hot vectors compared to input images, which is reflected in the lower complexity of the distribution $p(\theta)$ over parameters $\theta$, as predicted by a model, relative to that of $p(x)$. 

In this work, we develop a general model capable of \textit{learning the distribution} $p(\theta)$ over predicted parameters (the ``prediction distribution'') to enable semi-supervised learning for tasks of arbitrary output complexity. Thus, we consider only other SSL methods that rely on assumed properties of the prediction distribution. 
Existing SSL approaches for $K$-class classification pertaining to $p(\theta)$, such as entropy minimisation \cite{grandvalet2005semi} and mutual exclusivity \cite{sajjadi2016mutual, xu2018semantic}, have been shown to rely on strong prior assumptions about $p(\theta)$ (see Section \ref{sec:ssl}). However, the analytic form of $p(\theta)$ depends on the data, so prior knowledge of the prediction distribution may not be available and strong a priori assumptions, e.g. minimising entropy, may be inappropriate. Further, in certain cases, the distribution $p(\theta)$ may be complex and not possible to define analytically, e.g. image-to-image translation.

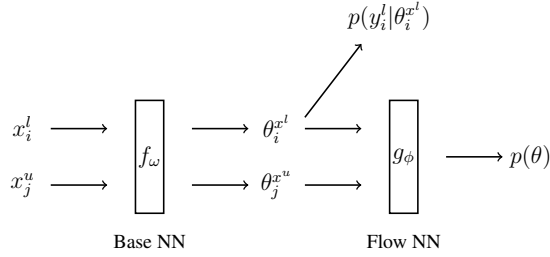
\begin{figure}[!t]
\centering
\resizebox{7.5cm}{!}{
\begin{tikzpicture}
\node[black] at (0, 0) {\large$x^l_i$};
\node[black] at (0, -1) {\large$x^u_j$};
\draw [thick, ->] (0.5, 0) -- (1.5, 0);
\draw [thick, ->] (0.5, -1) -- (1.5, -1);
\draw [thick] (2,0.5) -- (2.5,0.5) -- (2.5,-1.5) -- (2,-1.5) -- cycle;
\node[black] at (2.25, -0.5) {\large$f_\omega$};
\draw [thick, ->] (3, 0) -- (4, 0);
\draw [thick, ->] (3, -1) -- (4, -1);
\node[black] at (4.5, 0) {\large$\theta^{x^l}_i$};
\node[black] at (4.5, -1) {\large$\theta^{x^u}_j$};
\draw [thick, ->] (5, 0) -- (6, 0);
\draw [thick, ->] (5, -1) -- (6, -1);
\draw [thick] (6.5,0.5) -- (7,0.5) -- (7,-1.5) -- (6.5,-1.5) -- cycle;
\node[black] at (6.75, -0.5) {\large$g_\phi$};
\draw [thick, ->] (7.5, -0.5) -- (8.5, -0.5);
\node[black] at (9, -0.5) {\large$p(\theta)$};

\draw [thick, ->] (5, 0.2) -- (6, 1.5);
\node[black] at (6.5, 2) {\large$p(y^l_i | \theta^{x^l}_i)$};

\node[black] at (2.25, -2) {Base NN};
\node[black] at (6.75, -2) {Flow NN};

\end{tikzpicture}
}
\caption{Illustration of the LP method. The flow NN $g_\phi$ learns to approximate the prediction distribution $p(\theta)$ from labelled predictions $\theta^{x_l}_i$. The base NN $f_\omega$ learns to optimise the distribution $p(y^l_i | \theta^{x_l}_i)$ for labelled data and  $g_\phi(\theta^{x_u}_j)$ for unlabelled data. 
\label{fig:model}
\vspace{-10pt}
} 

\end{figure}

In looking to develop a method for SSL that relies on $p(\theta)$ without hard coding prior assumptions, we note that an empirical distribution over predictions $\{\Tilde\theta^x\}_{x\in\mX}$ should be indifferent to whether samples $x\!\in\!\mX$ are labelled or unlabelled. Thus, a posterior distribution over (sufficiently accurate) model predictions for labelled data can serve as a prior distribution over predictions of $\theta$ on unlabelled data. Exploiting recent advances in generative modelling, we propose using normalising flows to learn the distribution over labelled predictions. Following from Equation \ref{eq:SSL_new}, the full loss function for our \textit{learned prior (LP)} model is:
\begin{equation}
    \ell^\text{\,LP} \!=\! 
    - \frac{1}{N_l}\sum_{i=1}^{N_l}\log  p(y^l_i|\Tilde\theta^{x^l_i})
     - \frac{1}{N_l}\sum_{i=1}^{N_l} \log g_\phi(\Tilde\theta^{x^l_i})
    - \frac{1}{N_u}\sum_{j=1}^{N_u} \log g_\phi(\Tilde\theta^{x^u_j}),
    \label{eq:flow}
\end{equation}
where $\tilde\theta^{x} \!\approx\!f_{\omega}(x)$ are the outputs of a base neural network $f_\omega$ with parameters $\omega$ and $g_{\phi}(\tilde\theta^{x}) \!\approx\! p(\Tilde\theta^{x} | \alpha)$ is the probability of a prediction $\tilde\theta^{x}$ estimated by a normalising flow $g_\phi$ (with neural network parameters $\phi$). We minimise the first and last term of Equation \ref{eq:flow} with respect to $\omega$ and the second term with respect to $\phi$.

Intuitively, the first term of Equation \ref{eq:flow} encourages labelled predictions $\Tilde\theta^{x^l}$, output by the base neural network $f_\omega$, to approximate the true parameters $\theta^{x^l}$ of $p(y^l|x^l;\theta)$. Labelled predictions $\Tilde\theta^{x^l}$ are (i) used to update the base NN parameters $\omega$ (first term of Equation \ref{eq:flow}) and (ii) input to the flow $g_\phi(\Tilde\theta^{x^l})$ to update the flow NN parameters $\phi$ to learn to approximate $p(\theta|\alpha)$ with $g_\phi(\theta)$ (second term of Equation \ref{eq:flow}). Unlabelled predictions $\Tilde\theta^{x^u}$ are input to the flow to obtain their probability estimates under the flow NN's approximate distribution $g_\phi(\theta)$ (last term of Equation \ref{eq:flow}, corresponding to $\ell^\text{\,LP}_u$), by which $\omega$ is updated to refine predictions $\Tilde\theta^{x^u}$ such that they become more likely (i.e. move towards higher probability density) under $g_\phi(\theta)$. Due to its generality, Equation \ref{eq:flow} is applicable to a broad range of tasks where the prediction distribution can be learned;  and the flow can be considered a ``universally relevant prior'' \cite{grandvalet2005semi} for predictions on the unlabelled data.

Since $y$ is a deterministic function of $x$ for all tasks considered in this work, the prediction distribution is assumed \textit{equivalent to the label distribution}, i.e. $p(\theta) \!=\! p(y)$ \cite{allen2020probabilistic}. In practice, to minimise the influence of error in predictions introduced by the base neural network on the distribution learned by the flow, instead of learning the prior for the unlabelled data from predictions $\tilde\theta$, we learn it from the labels $y$ directly. We leave tasks with a stochastic mapping $x \to y$, where prediction and label distribution are not equivalent and the prior must be learned from the predictions, to future work.  

Note that the paradigm for learning the prediction distribution is not dependent on a particular density estimation method,  i.e. the flow neural network is a separately defined module from the base neural network and can be readily substituted for another density estimation method or some other means of defining the prior altogether, e.g. in terms of logical rules \cite{allen2020probabilistic}.

\textbf{Algorithm description}\hspace{0.1in}A batch of data contains an equal number $N$ of labelled $\{x^l_i, y^l_i\}_{i\in 1...N}$ and unlabelled samples $\{x^u_j\}_{j\in 1...N}$. Each input $x \in \{x^l, x^u\}$ is transformed through a function defined by the base neural network $f_{\omega}$ to obtain a prediction $\tilde \theta^{x} \!=\! f_{\omega}(x)$ of ground truth parameter $\theta^x$. Separately, each $\theta^{x^l}$ (equivalent to $y^l$ when $y|x$ is deterministic) is input to the normalising flow to learn $g_\phi(\theta)$. Labelled loss $\ell_l^\text{\,LP}(\tilde\theta^{x^l}\!, y^l)$ is computed for each $x^l$ and unlabelled loss $\ell_u^\text{\,LP}(\tilde \theta^{x^u}) \!=\! \!-\! \log g_{\phi}(\tilde \theta^{x^u})$ for each $x^u$. An illustration of the described method is shown in Figure \ref{fig:model}.

\section{Experiments} \label{experiments}

To demonstrate the generality of our approach, we evaluate it on a range of different tasks in order of the complexity of the labels: image classification, attribute prediction, and image-to-image translation. In the case of image classification and attribute prediction, $y|x$ is deterministic (i.e. each image is only paired with one label) and $p(\theta)$ can be defined analytically as a discrete distribution \cite{grandvalet2005semi}. Thus, although these tasks are not natural settings for a flow which is a continuous function approximator, they provide a useful proof of concept in which we can compare to methods that make use of the known distribution $p(\theta)$, before turning to more complex tasks. We make the  PyTorch code for all experiments publicly available.\footnote{\texttt{https://github.com/ibalazevic/lp-ssl}} Hyperparameter selection for both the base and flow NNs, as well as the model specifics of the flow NN, are described in Appendix \ref{hyperparams}. 

\vspace{-0.1cm}
\subsection{Datasets}
\vspace{-0.1cm}

\textbf{Image classification}\hspace{0.1in}For image classification, we use the standard SSL setup on two widely known datasets: SVHN \cite{netzer2011reading} (1000 labelled samples) and CIFAR-10 \cite{krizhevsky2009learning} (4000 labelled samples), containing 32x32 pixel images from 10 different classes. Standard data augmentation techniques are applied, such as random crops and horizontal flips for CIFAR-10 and translations for SVHN.

\textbf{Attribute prediction}\hspace{0.1in} We use the Animals with Attributes 2 (AwA2) dataset \cite{xian2018zero} for attribute prediction, containing 37,322 images of animals paired with an 85-dimensional binary attribute vector belonging to one of 50 classes, indicating presence or absence of a particular feature. We create a random split of 30,000 images in the training set, of which 4000 are used as labelled, and 1000 and 6,322 images in the validation and test sets respectively. Each image is resized to 64x64 pixels and random crops and horizontal flips are applied.

\textbf{Image-to-image translation}\hspace{0.1in} For the image-to-image translation experiments, we use two well known datasets, edges2shoes \cite{isola2017image} and Look Into Person (LIP) \cite{gong2017look}. edges2shoes contains 50,025 pairs of hand-drawn images and corresponding photos of shoes from the UT Zappos50K dataset \cite{yu2014fine}, split into 49,825 and 200 test images. LIP dataset contains 30,462 pairs of photos of people and semantically segmented image labels belonging to 19 body parts and clothing items, divided into 30,462 training images and 10,000 test images. Of the training images for each dataset, we use 10,000 as labelled. Each image is resized to 64x64 pixels and random horizontal flips are applied.

\subsection{Image classification}

\textbf{Experimental setup}\hspace{0.1in} Images are assigned to one of $K$ mutually exclusive classes. Following \cite{oliver2018realistic}, we choose a Wide ResNet \cite{zagoruyko2016wide} (more specifically “WRN-28-2” with depth 28 and width 2) as the base neural network for all image classification experiments. We do not change the standard model specification for WRN-28-2, so we refer the reader to \cite{zagoruyko2016wide} for model specifics. Softmax cross entropy loss is used as the labelled loss component. For fair comparison to existing methods relating to $p(\theta)$, we re-implement entropy minimisation \cite{grandvalet2005semi} and semantic loss \cite{xu2018semantic} in the same framework. As the output distribution is discrete, we add noise to outputs passed to the flow by sampling from a per-class continuous Dirichlet prior $Dir(\alpha)$ with $\alpha \!\in\! (\sR^+)^K; \alpha_k\!=\!120, \alpha_{k'\neq k}\!=\!1.1$ (corresponding to $\theta \!\in\! \Delta^K \!\subseteq\!\sR^K; \theta_k\!=\!1, \theta_{k'\neq k}\!=\!0$). We also experimented with adding random Gaussian noise $\gN(\mu, \sigma^2), \mu=0, \sigma^2=0.005$ to each dimension with similar classification performance. The unlabelled weighting parameter is set to $\lambda\!=\!0.1$ for minimum entropy and semantic loss and to $\lambda\!=\!0.01$ for the LP model on CIFAR-10 and $\lambda\!=\!0.005$ on SVHN. All models are trained for 200 epochs (including the flow NN) with batch size 256. 

\textbf{Classification accuracy}\hspace{0.1in} To assess if the prediction distribution learned by the LP model is effective for semi-supervised learning, we evaluate all models on the standard CIFAR-10 and SVHN datasets. From Table \ref{table:cifar}, we see that the LP model, without embedding strong prior assumptions on the shape of the output distribution, performs comparably to the analytically defined methods on both datasets.

\begin{table}[!htbp]
    \caption{Accuracy on the CIFAR-10 ($N_l\!=\!4000$), SVHN ($N_l\!=\!1000$) and AwA2 ($N_l\!=\!4000$) datasets. Top row shows supervised performance using the whole dataset as labelled. The remaining rows show SSL performance for different types of priors (entropy minimisation \cite{grandvalet2005semi}, semantic loss \cite{xu2018semantic} and the LP model). Each result is reported as a mean and standard error over 5 independent runs.}
	\centering
    \resizebox{10cm}{!}{
    \begin{tabular}{llccc}
    \toprule 
    & $\ell_u$ & \multicolumn{1}{c}{CIFAR-10} & \multicolumn{1}{c}{SVHN} & \multicolumn{1}{c}{AwA2}\\
    \midrule
    all & $-$ & $94.64 \pm 0.08$ & $97.10 \pm 0.04$ & $63.74 \pm 0.21$\\
    \midrule
    \multirow{4}{*}{$N_l \!=\! 4\text{k}/1\text{k}$} & $-$ & $82.90 \pm 0.19$ & $86.53 \pm  0.24$ & $36.81 \pm 0.31$\\[0.05cm]
    & $\ell_u^{\text{\,MinEnt}}$ \cite{grandvalet2005semi} & $84.62 \pm 0.20$ & $91.03 \pm 0.12$ & $40.86 \pm 0.42$\\[0.05cm]
    & $\ell_u^{\text{\,Sem}}$ \cite{xu2018semantic} & $84.84 \pm 0.07$ & $90.68 \pm 0.07$ & $50.45 \pm 0.64$\\[0.05cm]
    & $\ell_u^{\text{\,LP}}$ (ours) & $84.37 \pm 0.08$ & $90.01 \pm 0.31$ & $47.54 \pm 0.42$\\
    \bottomrule
    \end{tabular}
    }
     \label{table:cifar}
 \end{table}
 
 \begin{figure}[!htbp]
 \vspace{-0.7cm}
\centering
\subfloat[Two vertices of $\Delta^{10}$.]{\centering\includegraphics[width=.35\linewidth]{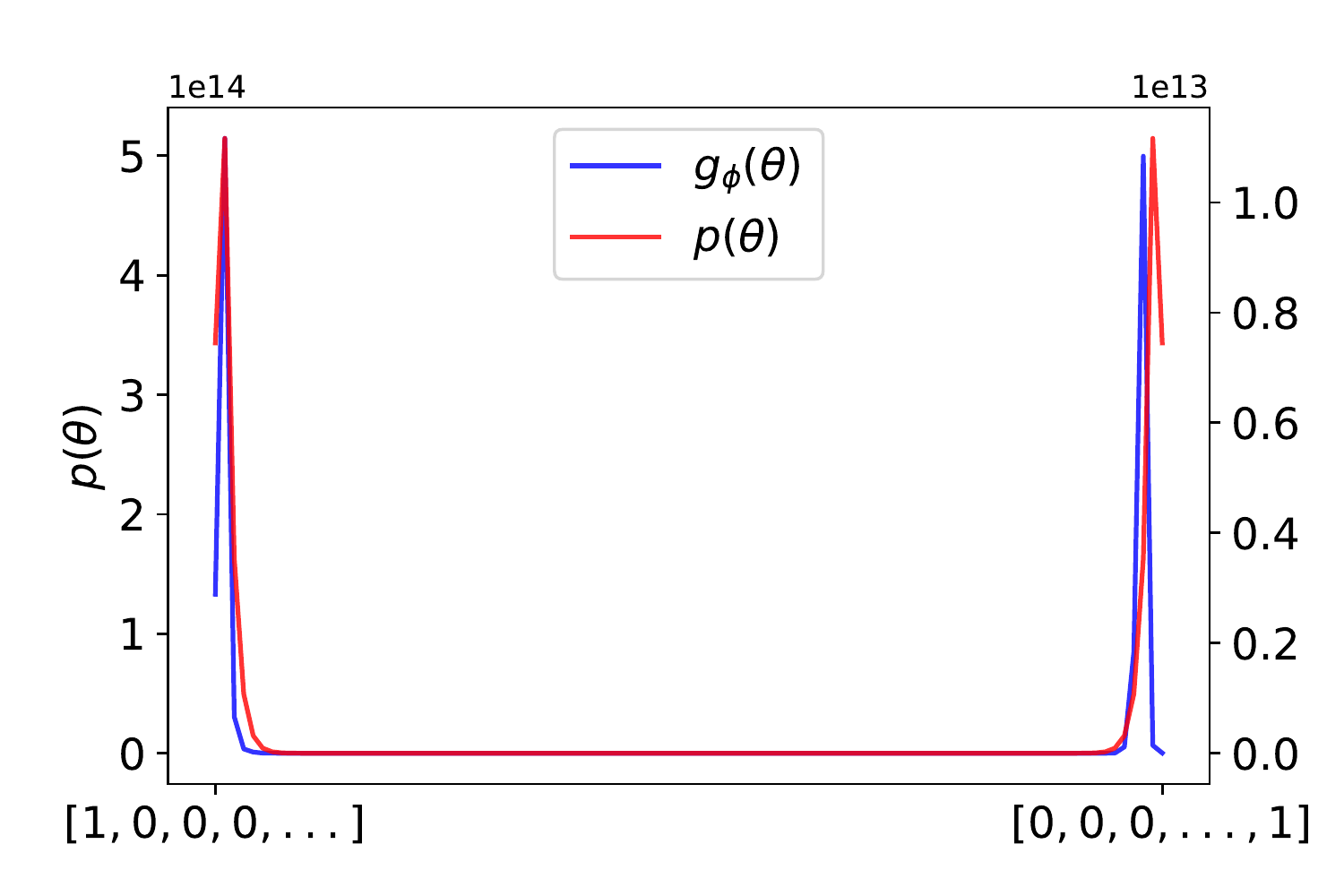}}\qquad
\subfloat[A vertex and a point on $\Delta^{10}$.]{\centering\includegraphics[width=.35\linewidth]{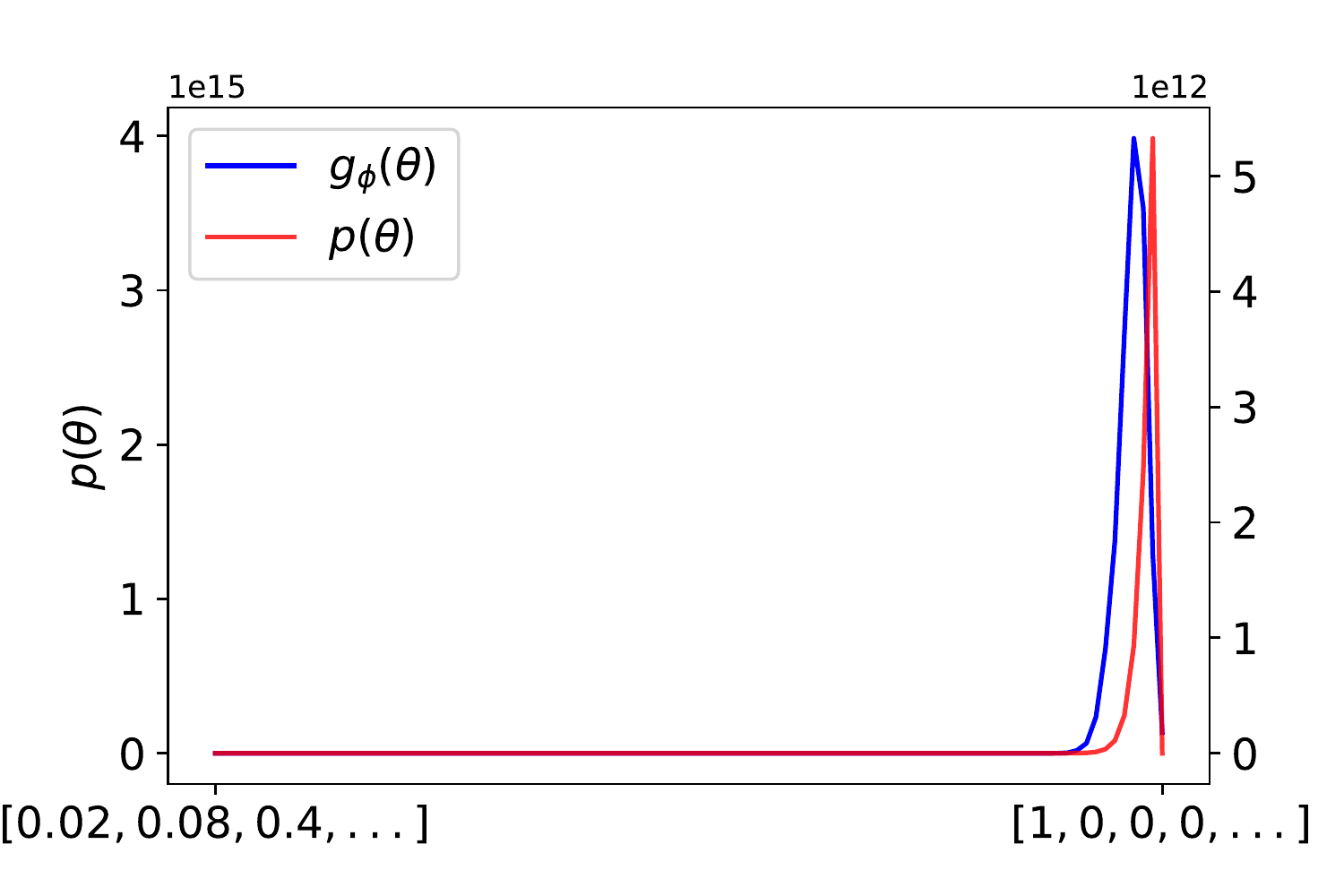}}
\caption{Slices of the analytically defined mixture of Dirichlets $p(\theta)$ vs distribution learned by the flow $g_\phi(\theta)$ evaluated at points on the 10-dimensional simplex $\Delta^{10}$. 
\vspace{-10pt}}
\label{fig:analysis}
\end{figure}

\textbf{Analysis of the learned distribution}\hspace{0.1in}Normalising flows are most commonly trained on images, where the distributions learned are typically smooth and continuous. To verify that the flow has learned to approximate the true discrete prediction distribution, we compute probabilities assigned by the flow between (i) two vertices of the 10-dimensional simplex $\Delta^{10}$ and (ii) a vertex and a random point on the simplex. Figure \ref{fig:analysis} shows that the distribution learned by the flow closely approximates the analytically defined mixture of Dirichlets.

\subsection{Attribute prediction}

\textbf{Experimental setup}\hspace{0.1in}In the attribute prediction task, an image is labelled with a set of binary attributes, indicating the presence or absence of a particular feature.
We use the same base neural network architecture as for image classification (WRN-28-2), with the output dimensionality of the final layer as the only difference to the original model (85 for AwA2 vs 10 for CIFAR-10 and SVHN).
Binary cross entropy loss is used as the labelled loss component. We re-implement the unlabelled loss components of semantic loss  and entropy minimisation. The discrete outputs are in this case relaxed by sampling from a $K$-dimensional Beta prior, with $Beta(\alpha\!=\!120, \beta\!=\!1.1)$ if $y_k\!=\!1$ and $Beta(\alpha\!=\!1.1, \beta\!=\!120)$ if $y_k\!=\!0$. As with the image classification experiments, we tested adding random Gaussian noise $\gN(\mu, \sigma^2), \mu=0, \sigma^2=0.005$ to each dimension with similar performance. All models are trained for 200 epochs with batch size 256. The unlabelled weighting parameter is set to $\lambda\!=\!0.005$
for minimum entropy and $\lambda\!=\!0.003$ for the LP model and semantic loss. 

\textbf{Attribute prediction accuracy}\hspace{0.1in} We compare all models in the SSL scenario on the AwA2 dataset. A prediction is considered correct only if it exactly matches the label, i.e. ${\tilde\theta}_k \!=\! y_k, \forall k \!\in\! K$. Table \ref{table:cifar} shows that adding the prior learned by the flow improves accuracy over the supervised baseline by $\sim\!10.7\%$. Minimum entropy cannot model the global class structure and unsurprisingly underperforms compared to semantic loss and the LP model. Semantic loss performance represents an approximate upper bound on SSL performance of prediction-based models, in that it is a smooth relaxation of the true prior based on \textit{known information} of the 50 class labels within the $2^{85}$-dimensional hypercube $\sH^{85}$. However, such knowledge may not be available a priori, or determining all possible labels within $\sH^{85}$ may be intractable. In contrast, the LP model requires no prior knowledge of the task.

\subsection{Image-to-image translation}

\textbf{Experimental setup}\hspace{0.1in}Image-to-image translation refers to a family of computer vision problems in which a mapping is learned between input and output images, e.g. aerial photos to maps and natural images to their segmentations. We consider only deterministic mappings. 
We evaluate the proposed LP model against the  state-of-the-art supervised image-to-image translation model pix2pix \cite{isola2017image}, which uses per pixel L1 loss computed on the output of a generator network and an adversarial loss produced by a discriminator network. We use the generator network of \cite{zhu2017unpaired} as a base network in all image-to-image translation experiments, and the discriminator network of \cite{isola2017image}. We adopt the standard specification of the two networks, so we refer to \cite{zhu2017unpaired} and \cite{isola2017image} for model specifics. In the semi-supervised setting, we use the per-pixel L1 loss\footnote{Note that in semantic segmentation, softmax cross entropy loss is more appropriate. However, this entails a very high output dimensionality $N_c\!\times\!64\!\times\!64$ (where $N_c\!=\!19$ is the number of classes) compared to the number of labelled samples $N_l \!=\! 10,000$, which is too challenging for the flow at present.} as a supervised loss component for all models; and compare our LP loss with the adversarial (``adv'') pix2pix loss as an unsupervised component. All models are trained for 20 epochs on edges2shoes with $\lambda\!=\!1$ for the adversarial and the LP loss and 100 epochs on LIP with $\lambda\!=\!0.1$ for the adversarial and $\lambda\!=\!0.05$ for the LP loss with batch size 128.

\begin{figure}[!htbp]
\vspace{-0.3cm}
\centering
\rotatebox[origin=l]{0}{$x$}\quad
\subfloat{\centering \hspace{0.5cm}
  \includegraphics[width=.85\linewidth]{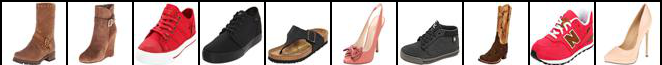}}
  \vspace{-0.39cm}
  
  \rotatebox[origin=l]{0}{$\tilde\theta_\text{sup}$}\quad
  \subfloat{\centering \hspace{0.17cm}
  \includegraphics[width=.85\linewidth]{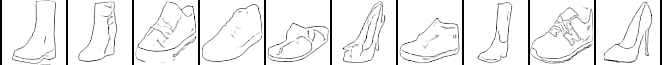}}
  \vspace{-0.5cm}
    
  \rotatebox[origin=l]{0}{$\tilde\theta_\text{adv\,\,}$}\quad
  \subfloat{\centering \hspace{0.08cm}
  \includegraphics[width=.85\linewidth]{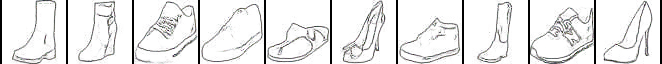}}
  \vspace{-0.45cm}
    
  \rotatebox[origin=l]{0}{$\tilde\theta_\text{LP}$}\quad
  \subfloat{\centering \hspace{0.23cm}
  \includegraphics[width=.85\linewidth]{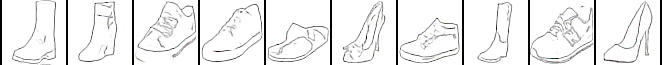}}
  \vspace{-0.44cm}

  \rotatebox[origin=l]{0}{$y$}\quad
  \subfloat{\centering \hspace{0.51cm}
  \includegraphics[width=.85\linewidth]{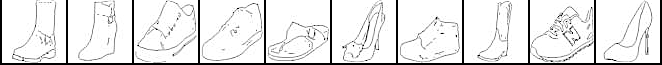}}

\caption{A random sample from the edges2shoes dataset with $N_l\!=\!10,000$. First and last row contain input and target images respectively.  $\smash{\tilde\theta_\text{sup}}$ denotes predictions with the labelled L1 loss only, $\smash{\tilde\theta_\text{adv}}$ those with the added adversarial loss and $\smash{\tilde\theta_\text{LP}}$ those with the learned prior loss (ours). It can be seen that images produced by the LP model are often more detailed than those produced by other approaches, in some cases even more so than the target images (e.g. columns 4 and 7).}
\label{fig:e2s}
\end{figure}

\begin{figure}[!htbp]
\vspace{-0.7cm}
\centering
\subfloat[$x$]{\centering
  \includegraphics[width=.14\linewidth]{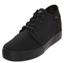}}\hspace{0.5cm}
  \subfloat[$\tilde\theta_\text{sup}$]{\centering 
  \includegraphics[width=.14\linewidth]{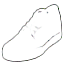}}\hspace{0.5cm}
  \subfloat[$\tilde\theta_\text{adv}$\label{fig:shoe_adv}]{\centering 
  \includegraphics[width=.14\linewidth]{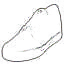}}\hspace{0.5cm}
  \subfloat[$\tilde\theta_\text{LP}$]{\centering 
  \includegraphics[width=.14\linewidth]{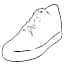}}\hspace{0.5cm}
  \subfloat[$y$\label{fig:shoe_gt}]{\centering 
  \includegraphics[width=.14\linewidth]{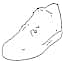}}

\caption{A closer look at an example where the learned prior loss improves SSL prediction. Prediction by the supervised only model $\smash{\tilde\theta_\text{sup}}$ approximately matches the ground truth $y$. Prediction with added adversarial loss $\smash{\tilde\theta_\text{adv}}$ appears noisy, while that with the learned prior (LP) loss $\smash{\tilde\theta_\text{LP}}$ appears the most detailed compared to other models and even the ground truth.}
\label{fig:shoe}
\end{figure}

\begin{figure}[!htbp]
\vspace{-0.5cm}
\centering
\rotatebox[origin=l]{0}{$x$}\quad
\subfloat{\centering \hspace{0.495cm}
  \includegraphics[width=.85\linewidth]{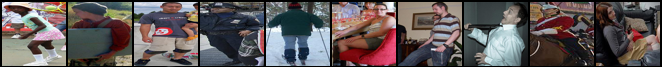}}
  \vspace{-0.39cm}
  
  \rotatebox[origin=l]{0}{$\tilde\theta_\text{sup}$}\quad
  \subfloat{\centering \hspace{0.17cm}
  \includegraphics[width=.85\linewidth]{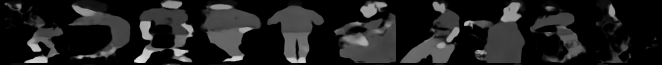}}
  \vspace{-0.5cm}
  
  \rotatebox[origin=l]{0}{$\tilde\theta_\text{adv}$}\quad
  \subfloat{\centering \hspace{0.155cm}
  \includegraphics[width=.85\linewidth]{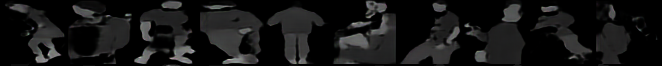}}
  \vspace{-0.47cm}
    
  \rotatebox[origin=l]{0}{$\tilde\theta_\text{LP}$}\quad
  \subfloat{\centering \hspace{0.225cm}
  \includegraphics[width=.85\linewidth]{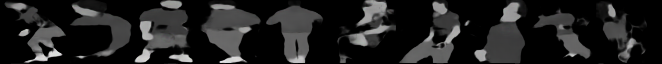}}
  \vspace{-0.47cm}

  \rotatebox[origin=l]{0}{$y$}\quad
  \subfloat{\centering \hspace{0.51cm}
  \includegraphics[width=.85\linewidth]{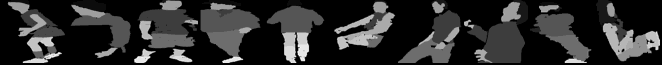}}

\caption{A random sample from the LIP dataset with $N_l\!=\!10,000$. First and last row contain input and target images respectively. $\smash{\tilde\theta_\text{sup}}$ denotes predictions with the labelled L1 loss only, $\smash{\tilde\theta_\text{adv}}$ those with the added adversarial loss and $\smash{\tilde\theta_\text{LP}}$ those with the learned prior loss (ours). Predictions by the LP model appear more coherent and sharper than those of other approaches (e.g. columns 1, 2, 3 and 10).
\vspace{-22pt}
}
\label{fig:lip}
\end{figure}

\textbf{Image-to-image translation results}\hspace{0.1in}Figures \ref{fig:e2s} and \ref{fig:lip} show a random sample of predicted images for all models. Predictions by the LP model $\tilde\theta_\text{LP}$ on edges2shoes (Figure \ref{fig:e2s}) often contain more detail than the ground truth images (e.g. shoelaces in columns 4 and 7, shown more closely in Figure \ref{fig:shoe}). This is perhaps because the labels for some training set images may contain more detail than others, whereas a prediction model learns a deterministic function that may be more consistent across images. Even though predictions with the added adversarial loss $\tilde\theta_\text{adv}$ are in some cases more detailed than those without, they appear to contain noise not present in the ground truth $y$ (Figure \ref{fig:shoe_adv} vs \ref{fig:shoe_gt}). On the LIP dataset (Figure \ref{fig:lip}), predictions by the LP model often appear sharper and more semantically coherent than those by other models (e.g. in columns 1 and 10, $\tilde\theta_\text{LP}$ most closely resemble human shapes; in column 3, the trousers in $\tilde\theta_\text{LP}$ are cohesive). Images predicted by the supervised model $\tilde\theta_\text{sup}$ appear in most cases more realistic than those with the added adversarial loss $\tilde\theta_\text{adv}$, matching the findings of the pix2pix authors on the semantic segmentation task in the supervised scenario \cite{isola2017image}.

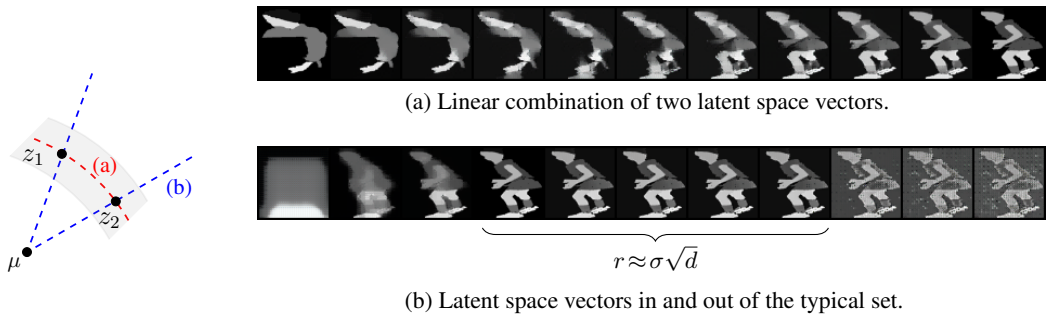
\begin{figure}[!htbp]
\centering
\begin{minipage}[b]{0.2\textwidth}
\resizebox{2.8cm}{!}{
\begin{tikzpicture}
 \draw[thick, red, dashed] (-1,-1) arc (30:70:3);
 \draw[thick, blue, dashed] (0,0) -- ++(30+180:3) -- +(70:3);
  \draw[thick, gray, opacity=0.1, name path=A] (-1.3,-1.3) arc (30:70:3);
  \draw[thick, gray, opacity=0.1, name path=B] (-0.7,-0.7) arc (30:70:3);
  \tikzfillbetween[of=A and B]{gray, opacity=0.1};
 
\node[black] at (-2.8, -1.7) {$\large\mu$};
\node[black] at (-2.5, 0) {\large$z_1$};
\node[black] at (-1.3, -1) {\large$z_2$};

\filldraw[black] (-2.6, -1.5) circle (2pt) node[anchor=west]{};
\filldraw[black] (-2.05, 0.05) circle (2pt) node[anchor=west]{};
\filldraw[black] (-1.2, -0.7) circle (2pt) node[anchor=west]{};

\node[red] at (-1.37, -0.15) {(a)};
\node[blue] at (-0.2, -0.5) {(b)};
\end{tikzpicture}
}
  \end{minipage}\hfill
\begin{minipage}[b]{0.75\textwidth}
\subfloat[Linear combination of two latent space vectors. \label{fig:comb}]{\centering\includegraphics[width=\linewidth]{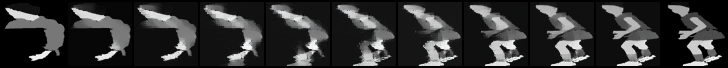}} \\
\subfloat[Latent space vectors in and out of the typical set.\label{fig:length}]{\centering

\begin{tikzpicture}

\node[anchor=south west,inner sep=0] (image) at (0,0)
{\includegraphics[width=\linewidth]{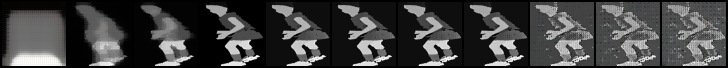}};

\draw[decorate, decoration={brace, amplitude=5pt}] (7.6,-0.1)--(3,-0.1);
\node[black] at (5.3, -0.5) {\small $r \!\approx\!\sigma\sqrt{d}$};
\end{tikzpicture}
}
\end{minipage}

 \caption{Samples from a flow created by (a) interpolating between two latent space vectors $z_1$ and $z_2$ along a band of similar density; and (b) scaling the length of $z_2$ to move from the mode through the typical set. $\mu$ indicates the mean of $p(z)$ and the shaded region denotes the typical set with $r \!\approx\!\sigma\sqrt{d}$.}
 \label{fig:analysis_img}
\end{figure}

\textbf{Analysis of the learned distribution}\hspace{0.1in}We take advantage of the invertibility of flows to analyse the learned prediction distribution.
It is well known that samples from a high-dimensional Gaussian distribution are not concentrated near the mode of the distribution, but within the \textit{typical set} \cite{betancourt2017conceptual}, a tight band with radius $r \!\approx\!\sigma\sqrt{d}$ (where $d$ is the dimensionality of the Gaussian). Figure \ref{fig:analysis_img} shows samples $\theta_i \!=\! g_\phi^{-1}(z_i)$ from the flow trained on the LIP dataset, corresponding to: (a) $z_i$ interpolated points between $z_1 \!=\!g_\phi(\theta_1)$ and $z_2 \!=\!g_\phi(\theta_2)$, for two test images $\theta_1, \theta_2$, following a band of similar probability density within the typical set; and (b) $z_i\!=\!z_2 c$ for scalar $c\!\in\![0,3]$, moving from the mode through the typical set. As expected, samples are the sharpest in the neighbourhood of the typical set.

\vspace{-0.2cm}
\section{Conclusion}\label{conclusion}
\vspace{-0.2cm}
Substantial progress has been made on semi-supervised learning for image classification, in large part from methods exploiting assumptions of the input distribution $p(x)$.
While image classification is an important and ubiquitous problem, many other tasks in the image domain and beyond can benefit from leveraging unlabelled data. 
Inspired by a recent theoretical framework for discriminative SSL, which focuses on the prediction distribution $p(\theta)$, we propose a general discriminative approach to SSL that can apply to tasks of arbitrary output distribution complexity. Our method uses normalising flows to learn the prediction distribution from labelled predictions, applying it simultaneously as a prior over the predictions on unlabelled data. We demonstrate the generality of the proposed approach on several tasks in the image domain: classification, attribute prediction and image-to-image translation. Having tested the concept in this familiar domain, in future work we intend on applying it to domains in which SSL methods tailored to images are less applicable.

\section*{Broader Impact}

In this work, we propose a general framework for semi-supervised learning in the image domain that is applicable to tasks with an arbitrary output distribution complexity, ranging from classification to attribute prediction and image-to-image translation. In contrast to existing approaches, which hard code a prior on the prediction distribution corresponding to assumptions specific to a particular task, we use normalising flows to learn the prediction distribution from labelled predictions and apply it as a prior on the unlabelled predictions.

The main benefit of the proposed method is its wide applicability to tasks of various prediction distribution complexity with limited labelled data. For example, obtaining depth maps from 2D images is expensive and learning a mapping from a real-world image to a depth map in an SSL manner may benefit tasks ranging from robotics to self driving cars and augmented reality. However, since the prediction distribution is learned rather than analytically defined, the proposed method is susceptible to learning and reinforcing biases in the provided label distribution, which could have implications for fairness in machine learning applications.

\section*{Acknowledgements}
We would like to thank Artur Bekasov and Conor Durkan for helpful discussion on normalising flows.

\bibliographystyle{plainnat}
\bibliography{neurips_2020}

\clearpage
\appendix

\section{Hyperparameters}\label{hyperparams}

\textbf{Distribution parameters}\hspace{0.1in}Parameters of Dirichlet and Beta distributions are chosen from $\alpha_k\!=\!\{60, 80, 100, 120, 140, 160\}, \alpha_{k'\neq k}\!=\!\{1.1, 1.5, 2.0\}$ and $\alpha\!=\!\{20, 40, 60, 80, 100, 120, 140\}, \beta\!=\!\{1.01, 1.1, 1.5\}$ respectively by random search and evaluated on the validation set accuracy. When experimenting with adding random Gaussian noise, the mean is always set to $\mu=0$ and variance is chosen from $\sigma^2 \in \{0.001, 0.005, 0.01, 0.05, 0.1\}$.

\textbf{Flow NN hyperparameters}\hspace{0.1in}For the classification and attribute prediction datasets, we used the neural spline flow implementation from \nolinkurl{https://github.com/karpathy/pytorch-normalizing-flows}. We fixed the number of bins to $K\!=\!8$, the tail bound to $B\!=\!3$, the number of hidden features to $16$ for all experiments and number of flow steps to 3. The flow is trained using the Adam optimiser with learning rate $10^{-3}$ and weight decay $10^{-5}$. For the image-to-image translation experiments, we used the original neural spline flow implementation provided by the authors \nolinkurl{https://github.com/bayesiains/nsf}. The following hyperparameters are used across all experiments: number of bins $K\!=\!8$, tail bound $B\!=\!1$, 64 hidden features and 5 flow steps across 3 levels. The flow is trained using Adam with learning rate $5\times10^{-4}$. The latent distribution $p(z)$ is set to standard normal in all experiments.

\textbf{Other hyperparameters}\hspace{0.1in}The unlabelled weighting parameter is chosen by random search on the validation set from $\lambda\!=\!\{0.001, 0.003, 0.005, 0.01, 0.05, 0.1, 0.5, 1\}$. For image classification and attribute prediction, we used $\ell_2$ regularisation of $5\times10^{-4}$ in all experiments.

\section{Additional image-to-image translation results}
\vspace{-0.3cm}

\begin{figure}[!htbp]
\centering
\rotatebox[origin=l]{0}{$x$}\quad
\subfloat{\centering \hspace{0.5cm}
  \includegraphics[width=.8\linewidth]{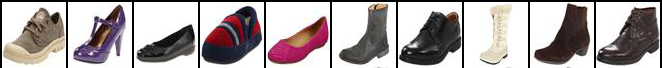}}
  \vspace{-0.39cm}
  
  \rotatebox[origin=l]{0}{$\tilde\theta_\text{sup}$}\quad
  \subfloat{\centering \hspace{0.175cm}
  \includegraphics[width=.8\linewidth]{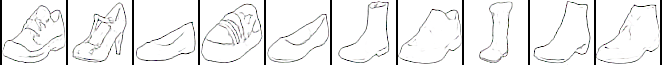}}
  \vspace{-0.5cm}
  
  \rotatebox[origin=l]{0}{$\tilde\theta_\text{adv}$}\quad
  \subfloat{\centering \hspace{0.16cm}
  \includegraphics[width=.8\linewidth]{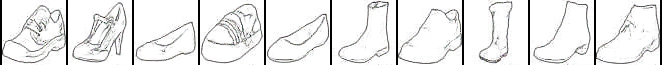}}
  \vspace{-0.45cm}
    
  \rotatebox[origin=l]{0}{$\tilde\theta_\text{LP}$}\quad
  \subfloat{\centering \hspace{0.23cm}
  \includegraphics[width=.8\linewidth]{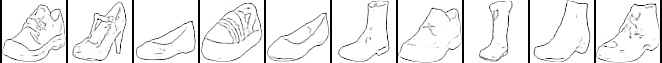}}
  \vspace{-0.44cm}

  \rotatebox[origin=l]{0}{$y$}\quad
  \subfloat{\centering \hspace{0.515cm}
  \includegraphics[width=.8\linewidth]{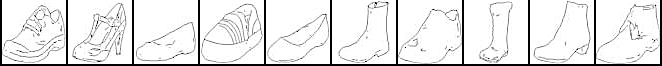}}
  
\caption{Additional edges2shoes SSL results.}
\label{fig:e2s_extra}
\end{figure}

\begin{figure}[!htbp]
\vspace{-0.7cm}
\centering
\rotatebox[origin=l]{0}{$x$}\quad
\subfloat{\centering \hspace{0.495cm}
  \includegraphics[width=.8\linewidth]{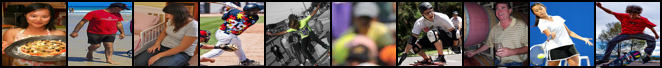}}
  \vspace{-0.39cm}
  
  \rotatebox[origin=l]{0}{$\tilde\theta_\text{sup}$}\quad
  \subfloat{\centering \hspace{0.177cm}
  \includegraphics[width=.8\linewidth]{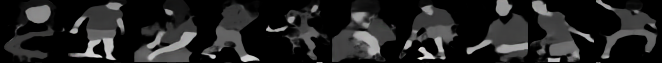}}
  \vspace{-0.5cm}
  
  \rotatebox[origin=l]{0}{$\tilde\theta_\text{adv}$}\quad
  \subfloat{\centering \hspace{0.16cm}
  \includegraphics[width=.8\linewidth]{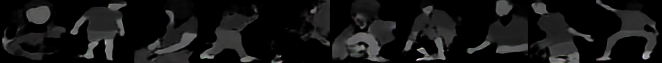}}
  \vspace{-0.5cm}
    
  \rotatebox[origin=l]{0}{$\tilde\theta_\text{LP}$}\quad
  \subfloat{\centering \hspace{0.23cm}
  \includegraphics[width=.8\linewidth]{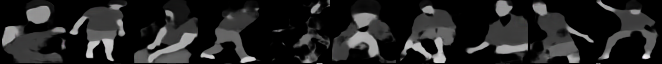}}
  \vspace{-0.47cm}

  \rotatebox[origin=l]{0}{$y$}\quad
  \subfloat{\centering \hspace{0.515cm}
  \includegraphics[width=.8\linewidth]{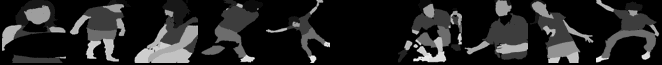}}
  
\caption{Additional LIP SSL results.}
\label{fig:lip_extra}
\end{figure}

\end{document}

%% file: math_commands.tex
\usepackage{amsmath,amsfonts,bm}

\def\eqref#1{equation~\ref{#1}}

\def\1{\bm{1}}

\def\mX{{\bm{X}}}
\def\mY{{\bm{Y}}}

\DeclareMathAlphabet{\mathsfit}{\encodingdefault}{\sfdefault}{m}{sl}
\SetMathAlphabet{\mathsfit}{bold}{\encodingdefault}{\sfdefault}{bx}{n}

\def\gN{{\mathcal{N}}}

\def\gX{{\mathcal{X}}}

\def\gZ{{\mathcal{Z}}}

\def\sH{{\mathbb{H}}}

\def\sR{{\mathbb{R}}}

\def\sY{{\mathbb{Y}}}

